\documentclass[a4paper]{article}
\usepackage{multirow}
\usepackage{INTERSPEECH2021}

\title{SynthASR: Unlocking Synthetic Data for Speech Recognition}
\name{Amin Fazel, Wei Yang, Yulan Liu, Roberto Barra-Chicote, Yixiong Meng, Roland Maas, Jasha Droppo}
\address{Alexa Speech, Amazon.com}
\email{\{aminfaze, wyanamz, lyulan, rchicote, myixiong, rmaas, drojasha\}@amazon.com}

\begin{document}

\maketitle

\begin{abstract}
End-to-end (E2E) automatic speech recognition (ASR) models
have recently demonstrated superior performance over the traditional hybrid ASR models. 
Training an E2E ASR model requires a large amount of data which is not only expensive 
but may also raise dependency on production data. 
At the same time, synthetic speech generated by the state-of-the-art text-to-speech (TTS)
engines has advanced to near-human naturalness.
In this work, we propose to utilize synthetic speech 
for ASR training (SynthASR) in applications where data is sparse or hard to get for ASR model training.
In addition, we apply continual learning with a novel multi-stage training strategy
to address catastrophic forgetting, achieved by
a mix of weighted multi-style training, data augmentation, 
encoder freezing, and parameter regularization.
In our experiments conducted on in-house datasets for a new application of recognizing
medication names, training ASR RNN-T models with synthetic audio via the proposed 
multi-stage training improved the recognition performance on new application by 
more than 65\%  relative, without degradation on existing general applications. 
Our observations show that SynthASR holds great promise in training 
the state-of-the-art large-scale E2E ASR models for new applications while reducing
the costs and dependency on production data.

\end{abstract}

\section{Introduction}
\label{sec:introduction}

End-to-end (E2E) designs, such as those based on connectionist temporal classification 
(CTC)~\cite{graves2006connectionist, pmlr-v32-graves14}, 
recurrent neural network transducer (RNN-T)~\cite{graves2012sequence}, 
and Listen, Attend and Spell (LAS)~\cite{Chan_2016}, have several advantages over the
older hybrid designs for automatic speech recognition (ASR) tasks.
These designs jointly optimize the model parameters to improve the accuracy
at text level, and they learn the tasks directly from data.
The highly integrated model structure in E2E designs reduces the overall model size and simplifies both 
training and inference, making it more attractive to on-device applications~\cite{He2019rnnt}.

To perform well in real applications, E2E ASR systems need to be trained on thousands of hours 
of transcribed speech data. One way to meet this data requirement is to learn directly 
from transcribed production data. 
To work with production data, one must be careful to handle
the data properly.
This includes the methods by which the
data is collected,
transferred, stored, accessed, and deleted. 
It also includes minimizing
the amount of human exposure to the data, such as when it
is transcribed.

The goal of this work is to reduce the reliance on human transcribed data by training ASR models on production-like 
data synthesized from text by text-to-speech (TTS) engines~\cite{Ueno2019, Rossenbach2020, Du2020, Zheng2021}. 
The contribution of this work comes in three folds. First, we validate the effectiveness of using TTS based 
synthetic data as a general approach to reduce reliance on transcribed data. Second, we study how to improve ASR models 
for new applications without relying on corresponding production data. Third, we propose a multi-stage training strategy, 
and demonstrate that continual learning via this strategy significantly improves ASR performance on the new applications 
without degradation on existing applications.

\section{Related  Work} 
Recent research has made significant progress in using synthetic speech data for ASR model training. 
In \cite{Mimura2018}, a TTS engine based on Tacotron-2 is used to synthesize audio for new vocabulary to teach an acoustic-to-word speech recognition model new words. In a follow-up work \cite{Ueno2019}, multi-speaker TTS is used to improve the acoustic diversity in synthetic data, where speaker embeddings are added to Tacotron-2 architecture. Later in~\cite{Rossenbach2020},  global style token (GST) based embeddings are introduced to modify a version of Tacotron-2 to further increase the acoustic diversity of synthetic data where GST is found superior to i-vector based embeddings. In addition, this work also demonstrates that adding TTS based synthetic data, LM approaches, and general data augmentation method SpecAugment~\cite{Park_2019} are mostly independent and complementary. In~\cite{Du2020}, E2E TTS with speaker presentations from a variational autoencoder (VAE) is explored to increase the acoustic diversity in low-resource data. All these previous works in~\cite{Ueno2019, Rossenbach2020, Du2020} have shown the benefit of increased acoustic diversity in synthetic data for ASR model improvement, especially in low resource scenarios. 

The works in~\cite{Mimura2018, Ueno2019, Zheng2021} have shown that E2E ASR models can learn new vocabularies from TTS based synthetic data, which is crucial for feature expansion of ASR into new applications. 
Recent work by \cite{Zheng2021} pointed out some practical challenges in such vocabulary expansion strategy in terms of learning to recognize new words without recognition degradation on already learned words. Authors have explored several strategies to address this challenge including combining real data with synthetic data with weighted sampling and applying different regularizations on each model components. 

Inspired by previous research, this work reduces the dependency of ASR model training on production data by using TTS synthetic data.
While a number of existing methods can achieve similar goals, each has their own limitations. For example, in semi-supervised learning (SSL)~\cite{Synnaeve2019, Ling2020}, where speech audio is transcribed by machine rather than human, audio signals are still required to be collected, transferred, and stored in the cloud to generate machine transcribed labels. In federated learning (FL)~\cite{McMahanMRA16,LinHM0D18}, where many devices collaboratively train a shared global model, comprehensive infrastructure updates are needed, which puts additional cost burdens on the customers for device upgrades in order to support new features. 

Acknowledging the importance of language model (LM) approaches as alternative methods to utilize text-only data for E2E ASR performance improvement~\cite{McDermottSV19, Variani2020, Meng2021}, this work focuses on the WER improvement from the first pass ASR model to provide LM approaches a better starting point. In addition, we applied both SpecAugment and classical signal processing based data augmentation methods on TTS synthetic data.

\section{Technical Approaches}
\label{sec:methods}

\subsection{Overview}

\begin{figure}[t]
	\centering
	\includegraphics[width=\linewidth]{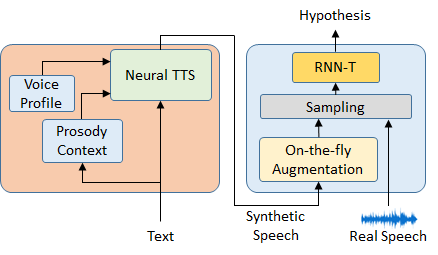}
	\caption{Diagram of the proposed SynthASR employing TTS synthetic data for ASR training.}
	\label{fig:overal-system}
\end{figure}

The schematic diagram of our proposed system is illustrated in Figure \ref{fig:overal-system}. 
It consists of a multi-context TTS engine to generate synthetic speech, and an RNN-T model for speech recognition. 
The model for TTS engine is trained and evaluated independently from the speech recognition model.
RNN-T models are trained in a multi-style training (MST)~\cite{Lippmann1987, Ko2017}. 
With MST, the training data are sampled from a combination of real speech recordings and TTS based synthetic speech audio.
The ratio between real recordings and synthetic audio seen during RNN-T training is optimized with sampling weights, as suggested by~\cite{Zheng2021}. 
This method well mixes the real and synthetic data in each batch so that the ASR model sees both data. 
In addition, data augmentation is applied at both audio level and feature level for RNN-T training.
For one production scale application covered in this work, instead of training RNN-T models from scratch, we adopted continual learning (CL)~\cite{Parisi2019,gepperth2016} where an established RNN-T model incrementally learns from data with new information without catastrophic forgetting via the proposed multi-stage training approach.

\subsection{Multi-context TTS}

\begin{figure}[t]
	\centering
	\includegraphics[width=\linewidth]{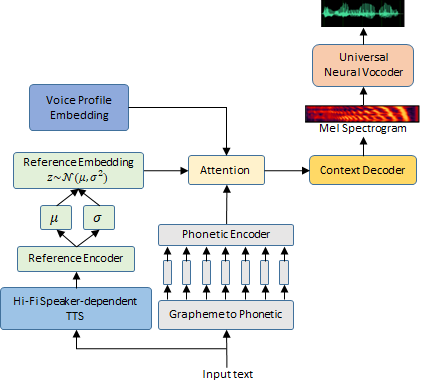}
	\caption{Architecture of proposed Multi-context TTS system.}
	\label{fig:tts}
\end{figure}

We use a multi-context TTS to generate clean synthetic speech with diverse speaker and prosody attributes (Figure~\ref{fig:tts}). 
This system consists of two main modules: a context generation module and a neural vocoder module. 
The context generation module is an attention-based sequence-to-sequence network~\cite{prateek2019news} that predicts a Mel-spectrogram given an input text.  
We control the speaker identity with voice profile embeddings, which introduces a bias in the training that makes the reference encoder to be speaker independent. 
At inference time we guide the TTS with a reference spectrogram generated with a high-fidelity speaker-dependent TTS system trained with more that 20 hours high quality data. 
This reference spectrogram provides a natural prosodic contour to the attention module. 
Since the reference encoder has been trained following a variational auto-encoder (VAE) approach, where we learn the posterior distributions over the prosody latent space given the reference spectrogram, we later can sample from the posteriors to generate different speech realizations of the same text. 
This one-to-many capability increases the inter- and intra- speaker variability, which is crucial for ASR training
as it increases the speaker diversity of synthetic data. 
The neural vocoder module consists of the architecture similar to the universal vocoder described in~\cite{rohnke2020parallel} but without any VAE reference. 
This Universal Neural Vocoder (UNV) was pretrained with more than 100 hours from more than 100 speakers in 27 languages from a proprietary database of paid voice actors. The UNV synthesizes speech audio out of the Mel-spectrograms generated by the first module.

A speaker verification system is used to produce the voice profile embeddings for the context generation module. The speaker verification system was trained on an internal Amazon data. This speaker verification model uses the architecture introduced in~\cite{Chung2020}. This pre-trained speaker verification system is used to provide voice profile embeddings in the Multi-context TTS system as in Figure~\ref{fig:tts}. We first generate the speaker embedding for each utterance of training data in TTS system, then we average all utterance-level embeddings from a given speaker as voice profile embedding for this speaker.

\subsection{Data augmentation}

Reverberation is introduced to TTS synthetic audio by convolving the audio with an acoustic impulse response (AIR) randomly selected from a pool of 10,000 available AIRs estimated from chirp signal measurements in real rooms.
Then an audio segment randomly sampled from an in-house dataset is added on top as background noise, with an SNR ranging from 10 to 20 dB.
To increase the acoustic diversity, for each on-the-fly audio corruption, there is a 60\% probability of reverberation and an independently 60\% probability of noise addition.
This provides a mixture of clean synthetic speech, synthetic speech with reverberation only, synthetic speech with background noise only, as well as synthetic speech with both reverberation and background noise.
In addition, SpecAugment \cite{Park_2019} is applied on the log Mel filter bank features generated from both synthetic data and real data for RNN-T training.
Two frequency masks are applied to each utterances and the maximal masked frequency percentage is 37.5\%. 
The maximal ratio of each time mask is 5\% of the utterance duration, 
and the number of time masks is proportional to utterance length, i.e. 5\% of the frame numbers 
without being larger than 10. 
Where masks applies, Gaussian noise is used with the same mean and variance from the masked values.

\subsection{ASR RNN-T model}
RNN-T is an E2E ASR model architecture suitable for streaming applications with proven competitive performance \cite{graves2012sequence,Li2020RNNT}. 
It consists of a transcription network (or encoder), a prediction network (or decoder), and a joint network. 
Targeting at streaming applications, we use LSTM~\cite{HochSchm97} layers for both the encoder and decoder. 
The encoder sequentially maps acoustic feature $\mathbf{x}$ to a high-level feature representation $\mathbf{h} = \operatorname{Enc}(\mathbf{x})$. 
The decoder network take previous labels in the sequence and generates a high level representation for next prediction. 
We use a feed-forward network for the joint network to combine the information from both the acoustic representations from encoder and the linguistic representations from decoder to make a joint prediction of next word piece.
The loss function for RNN-T parameter optimization is the negative log-posterior of the target label sequence $\mathbf{y}$:
\begin{equation}
\label{eq:rnnt-objective}
\mathcal{L} = -  \log P(\mathbf{y}| \mathbf{h}),
\end{equation}
where \( P(\mathbf{y}  | \mathbf{h}) = \sum_\mathbf{\hat{y}} P(\mathbf{\hat{y}} | \mathbf{h}) \), \( \mathbf{\hat{y}}  \in \mathcal{A}\). \(\mathcal{A}\) is the set of all possible alignments including blank labels between encoder feature representation \( \mathbf{h}\) and target label sequence \( \mathbf{y} \).
The Adam algorithm \cite{kingma2014adam} is used for the numerical optimization during training, with a learning rate scheduler in three stages: linear warm-up, hold and exponential decay.

\subsection{Continual learning with multi-stage training}
\label{sub-sec:CL}
Continual learning (CL)~\cite{Parisi2019,gepperth2016} imitates the incremental life-long learning ability in humans and animals by machine, through acquiring data over time, and learning, fine-tuning, and transferring knowledge over various tasks. 
CL reduces the cost from model training with repeated visit to a large amount of data, and it also reduces or even removes the dependency on previously harvested training data. 

In this work, we apply CL and let a general-purpose ASR model learn for a new application task of recognizing medication names, by
continually exposing the ASR model to the synthetic speech containing the medication names.  
To prevent catastrophic forgetting, we propose a multi-stage training strategy for continual learning,
including freezing the LSTM layers of encoder during fine-tuning with synthetic speech data and unfreezing all layers in later stages. 
When parameters are unfrozen, to ensure that the learned parameters do not deviate too much when trained with synthetic speech for a new application, an elastic penalty term is introduced to the RNN-T loss function that minimizes the distance between the new and previous model parameters:
\begin{equation}
\label{eq:l2-reqularization}
\mathcal{J} = \lambda \sum(\theta_{i,pre} -  \theta^*_{i,cur})^2,
\end{equation}
where $i$ indexes the RNN-T decoder parameters, $pre$ and $cur$ are the previous and current training stages, respectively, $*$ indicates the trainable parameters and $\lambda$ adjusts forgetting speed.

\section{Results}
\label{sec:results}

\begin{table}[t]
\caption{WERs on the LibriSpeech.}
\centering
\begin{tabular}{ c | c | c | c | c } 
\toprule
\multirow{2}{*}{\textbf{Model}} & \multicolumn{2}{c|}{\textbf{Training sets}} & \multicolumn{2}{c}{\textbf{WER}} \\
& real & synthetic & test-clean & test-other
 \\
\midrule
Benchmark & 960  & - & 7.29 & 17.41\\
\midrule
Baseline & 480  & -& 9.90 & 22.64\\ 
Baseline + TTS & 480  & 1150  & 8.66	& 20.78\\ 
\bottomrule
\end{tabular}
\label{tab:libri}
\end{table}

\begin{table*}[t]
\caption{NWERs for the application of recognition of medication name. The weight on the left in the column \textbf{Weights\%} is the percentage of samples in MST from real data and the weight on the right is the percentage of samples from synthetic data. \textbf{(R, S)} indicates whether real (R) or synthetic (S) audio is used during each stage of training.}
\centering
\begin{tabular}{ c | c | c | c | c | c | c | c | c} 
\toprule
\multirow{2}{*}{\textbf{Model}} & \textbf{Real Data} & \textbf{Synthetic Data} & \textbf{Weights\%} & \textbf{Freeze} & \textbf{Elastic} & \multicolumn{3}{c}{\textbf{NWER}} \\
& (hours) & (hours) & (R, S) & \textbf{Encoder} &  \textbf{Penalty} & Dev-Gen & Eval-Gen & Eval-Med \\
\midrule
Baseline & 50K & - & - & No & - & 100 & 100 & 100\\
\midrule
Stage 1 & 50K & 5k & (95, 5) & Yes & No & 100.99 & 101.32 & 21.06\\ 
Stage 2 & 50K & 5k & (98, 2) & No  & No & 100.54 & 100.98 & \textbf{13.70}\\ 
Stage 3 & 50K & -  & -  & No   & Yes & 100.14 & 100.89 & 21.47\\
Stage 4 & 50K & - & - & No & No & \textbf{99.18} & \textbf{99.72} & 34.56 \\ 
\bottomrule
\end{tabular}
\label{tab:med}
\end{table*}

\subsection{Experiments on LibriSpeech: synthetic speech for reducing transcribed data}
To evaluate the effectiveness of synthetic data, we first perform experiments on the LibriSpeech dataset \cite{Panayotov_libri}.
In these experiments, we use 64-dimensional Log-Mel-Frequency features extracted with 25ms window and 10ms shift. Each feature vector is stacked with 2 frames to the left and downsampled by a factor of 3 corresponding to a frame rate of 30msec. For the experiments on LibriSpeech data, we use six-layer LSTMs with 1024 units in the encoder. The decoder is a two-layer LSTM with 1024 units in each layer. The output size of the recognition encoder and the decoder is set to 640. We use a one-layer feed-forward joint network with 512 units and tanh activation. The output softmax layer dimensionality is 2501 which corresponds to blank label and 2500 word pieces: the most likely subword segmentation from a unigram word piece model \cite{kudo2018sentencepiece}. We use an adaptive variant of SpecAugment \cite{Park_2019}, as proposed in \cite{Park_2020}. We use Adam algorithm \cite{kingma2014adam} for optimization of all models, and the learning rate is scheduled based on warm-up, hold and decay strategy as proposed in \cite{Park_2019}. For each experimental run, we chose the best model based on its performance on the development set. 

LibriSpeech contains 960 hours of read speech data for training \cite{Panayotov_libri}.
As an ASR baseline with a limited amount of audio data, we assume that only half of the LibriSpeech training data is available,
i.e. 480 hours training data randomly selected from the all training data.
As shown in Table \ref{tab:libri}, an RNN-T model trained with 480 hours of data is 35.8\% relatively worse on test-clean 
when compared to an RNN-T model trained with all 960 hours training data.  
We then synthesize about 1150 hours audio data using our multi-context TTS system, and the input texts for TTS are the transcriptions of the missing 480 hours data. 
This TTS training set contains about 48k unique input texts, and each text utterance is synthesized with randomly selected 24 voice profiles from the total of 500 available voice profiles. 
We then trained an RNN-T model using MST combining with 480 hours real data and 1150 hours synthetic speech.
Compared to the baseline model trained with 480 hours real data alone, this improves the performance on test-clean by 12.5\% relative (Table \ref{tab:libri}).

\subsection{Experiments in real application: synthetic speech for medication names recognition}

We then expand a general-purpose ASR model for a new application of medication names recognition.
This new application has no available real recordings for ASR model training.
In the following set of experiments, we use synthetic speech to teach a general-purpose ASR model to recognize medication names.

We use a slightly different RNN-T architecture from previous LibriSpeech experiments. The encoder now consists of 5 LSTM layers and the softmax layer has an output vocabulary size of 4001 word pieces including the blank label.
Note that, the results in this section are reported in normalized WER (NWER) numbers, which is the regular word error rate (WER) divided by the WER of the baseline model on the same test set and then multiplied by 100. Therefore, the NWERs for baseline model are 100 for all test sets as shown in Table \ref{tab:med}. 
The baseline RNN-T model for general-purpose application is trained with a dataset of 50,000 hours real human utterances. 
This dataset is a collection of de-identified production utterances from voice-controlled far-field devices. A development set (Dev-Gen) and an evaluation set (Eval-Gen) are constructed with the same type of utterances, consisting of about 50 hours and 160 hours of data respectively. These two test sets are used to monitor the performance change on existing applications. 
To evaluate the performance on the new application of medication name recognition, we collected 8 hours of real human data containing utterances with medication names (Eval-Med).

To support feature expansion without real training data, we prepared 5000 hours TTS synthetic data. 
The texts are generated by randomly combining 150 unique text utterance
templates and 600 common medication names.
For each text utterance generated, we sample 32 voice profiles from 500 voice profiles. 
The generated clean synthetic audio is corrupted with noise and reverberation on-the-fly for RNN-T model training. 
With MST, the corrupted synthetic audio is combined with real recordings with configured data ratio parameters.
With continual learning, we start with the baseline general-purpose RNN-T model and fine-tune it so that the recognition performance on general test sets (Eval-Gen) maintains while the performance on the test set for medication names (Eval-Med) is largely improved.

One challenge in such continual learning is a balance between forgetting learned knowledge which causes degradation on Eval-Gen and learning new knowledge which strives to improve performance on Eval-Med.
We use multi-stage training to address this challenge, and our experiments concluded with 4 critical stages as shown in Table \ref{tab:med}.
The first stage is to fine-tune the baseline model with batches of data that contain 95\% real data and 5\% synthetic data where we fix the RNN-T encoder parameters. 
We train in this way for 57k iterations, during which process the learning rate decays from 5e-5 to 1e-5. This stages ramps up the parameters for decoder and joint network for the new application of  medication name recognition. 
In the second stage, we further fine-tune the model with both real and synthetic utterances with the portion of 98\% and 2\% in each batch and fixed learning rate of 1e-5. 
As shown in Table \ref{tab:med}, this improves the recognition performance on Eval-Med further from an NWER of 21.06 to 13.70. 
In our experiments we found  the first training stage with encoder freezing critical, and removing Stage 1 led to performance degradation.
At the same time, model at the end of stage 2 showed a small degradation compared to baseline on general test sets.

To recover the degradation, in the third and fourth stages, we only fine-tune models with real human speech. In the third stage, we include an elastic penalty as described in section \ref{sub-sec:CL} to minimize the deviation of the model parameters from the previous stage as the model has well learned medication names. We further fine-tune the model in the fourth stage without such regularization but with a small learning rate of 1e-5 to ensure the performance of the model doesn't degrade from the baseline. 
Note that in our experiments, we found the third stage critical and directly jumping from Stage 2 to Stage 4 led to worse results.
As shown in Table \ref{tab:med}, the final model from Stage 4 achieved slightly better performance on both general test sets compared to baseline model, and at the same time the recognition performance on Eval-Med is more than 65\% better than the baseline model.
This is achieved with 5k hours of synthetic training data without real recordings for medication names .

\section{Conclusions}

In this work, we propose to use synthetic speech for E2E ASR model training to reduce both data costs and production data reliance. In addition, we demonstrated how to effectively and incrementally improve ASR for a new application that customer audio data is not available at all. Using continual learning with our proposed multi-stage training, the best system relatively improves the WER on the new application by more than 65\% without compromise on the existing application. While the value of synthetic speech as ASR training data remains less than that of real speech, but synthetic speech shows great promise in training large-scale ASR for new applications.  

\section{Acknowledgements}
We would like to thank Charles Chang and Paul MacCabe for the high-level support of this research. 
In addition, we would like to acknowledge the Alexa Data Synthetic and Alexa ASR teams for
providing the infrastructure that this work has benefited from.

\bibliographystyle{IEEEtran}

\bibliography{refs}

\end{document}